\documentclass[conference]{IEEEtran}
\IEEEoverridecommandlockouts
\usepackage{cite}
\usepackage{amsmath,amssymb,amsfonts}
\usepackage{algorithmic}
\usepackage{graphicx}
\usepackage{textcomp}
\usepackage{xcolor}
\usepackage{multicol}
\usepackage{multirow}
\usepackage{caption}
\usepackage{hyperref}
\usepackage{xcolor}
\usepackage{soul}

\def\BibTeX{{\rm B\kern-.05em{\sc i\kern-.025em b}\kern-.08em
    T\kern-.1667em\lower.7ex\hbox{E}\kern-.125emX}}

\begin{document}

\title{Synthetic data enables faster annotation and robust segmentation for multi-object grasping in clutter\\
\thanks{Dongmyoung Lee, Wei Chen, and Nicolas Rojas are with the REDS Lab, Dyson School of Design Engineering, Imperial College London, 25 Exhibition Road, London, SW7 2DB, UK
{\tt\small (d.lee20, w.chen21, n.rojas)@imperial.ac.uk}
Wei Chen was supported in part by the China Scholarship Council and the Dyson School of Design Engineering, Imperial College London.}
}

\author{\IEEEauthorblockN{1\textsuperscript{st} Dongmyoung Lee}
\IEEEauthorblockA{\textit{Dyson School of Design Engineering} \\
\textit{Imperial College London}\\
London, United Kingdom \\
d.lee20@imperial.ac.uk}
\and
\IEEEauthorblockN{2\textsuperscript{nd} Wei Chen}
\IEEEauthorblockA{\textit{Dyson School of Design Engineering} \\
\textit{Imperial College London}\\
London, United Kingdom \\
w.chen21@imperial.ac.uk}
\and
\IEEEauthorblockN{3\textsuperscript{rd} Nicolas Rojas}
\IEEEauthorblockA{\textit{Dyson School of Design Engineering} \\
\textit{Imperial College London}\\
London, United Kingdom \\
n.rojas@imperial.ac.uk}
}

\maketitle

\begin{abstract}
Object recognition and object pose estimation in robotic grasping continue to be significant challenges, since building a labelled dataset can be time consuming and financially costly in terms of data collection and annotation.
In this work, we propose a synthetic data generation method that minimizes human intervention and makes downstream image segmentation algorithms more robust by combining a generated synthetic dataset with a smaller real-world dataset (hybrid dataset).
Annotation experiments show that the proposed synthetic scene generation can diminish labelling time dramatically.
RGB image segmentation is trained with hybrid dataset and combined with depth information to produce pixel-to-point correspondence of individual segmented objects. The object to grasp is then determined by the confidence score of the segmentation algorithm. Pick-and-place experiments demonstrate that segmentation trained on our hybrid dataset (98.9\%, 70\%) outperforms the real dataset and a publicly available dataset by (6.7\%, 18.8\%) and (2.8\%, 10\%) in terms of labelling and grasping success rate, respectively.
 Supplementary material is available at \href{https://sites.google.com/view/synthetic-dataset-generation}{https://sites.google.com/view/synthetic-dataset-generation}.
\end{abstract}


\section{Introduction}
In order for robots to be utilized in many real-world settings, they must first be able to interact with their environment. While this is a relatively simple task for restricted industrial settings, many automation settings involve objects which exhibit large variation in shape and appearance. This dramatically increases the complexity of the task, not only in terms of object identification and localization, but also in terms of grasping strategy. Further still, difficulty is compounded when multiple objects are present, introducing occlusion and depth variation to the task.
Typical approaches to multi-object grasping have used computer vision for object identification and localization, with many works utilizing deep learning (DL) to estimate object pose from RGB images~\cite{xiang2017posecnn}, handle object occlusion~\cite{wang2019densefusion}, and address generalized multi-object manipulation~\cite{luo2022skp, song2020hybridpose}. The vast majority of these have relied on prior knowledge of the object in the form of 3D CAD models or topological information, at least in their training stages. This allows the problem to be simplified to regressing and refining the pose of the known object in the scene, then a pre-defined grasping strategy can be utilized.

\begin{figure}[t!]
    \centering
    \vspace{2mm}
    \includegraphics[width=0.99\columnwidth]{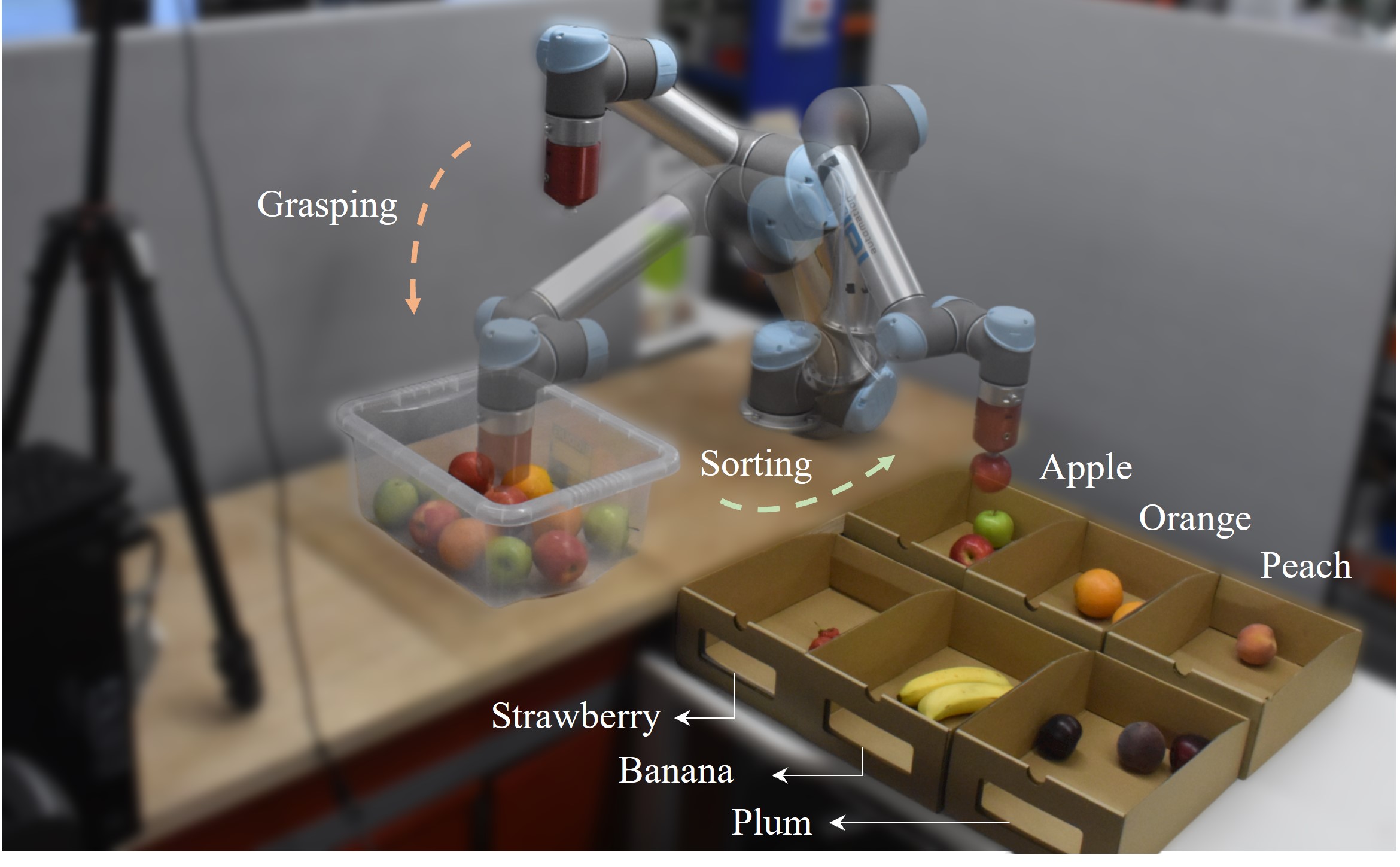}
    \caption{Robot learns to grasp multiple objects in clutter and sort them into target boxes with the proposed instance segmentation algorithm.}
    \label{fig:1}
    \vspace{-4mm}
\end{figure}

For these works to be applied to objects exhibiting a high degree of variation, this variation must be captured in the training setup. Unfortunately, this prior knowledge may be infeasible to obtain since it is difficult to encode in a 3D CAD model, as is the case for many organic objects such as fruits or vegetables.
Without prior knowledge of the 3D object model, images of the scene must first be segmented in order to identify at a fine-level which pixels belong to each object. From this, suitable grasping points can be extracted via stereographic reconstruction or from the depth channel of an RGB-D camera.
Although training a deep learning model to segment images is considerably easier than capturing object variation in 3D models, it still necessitates a considerable size and diversity of training data. Preparing labelled datasets for image segmentation is demanding. First, it is difficult to gather raw images with a wide range of settings and circumstances. Furthermore, the procedure of obtaining pixel-level labels makes data preparation even more expensive. To alleviate these issues, some works have utilized synthetic data generation to automatically produce and annotate usable training data~\cite{nguyen2018transferring, james2019sim}, however a sim-to-real gap limits performance. Therefore a hybrid approach, combining a mixture of synthetic and real training data, may be a fruitful avenue for research.

In this work, we consider the problem of fruit grasping (summarized in Fig.~\ref{fig:1}). With high intra-class variation, fruits are difficult to represent as 3D models, and incur a limiting cost when producing a suitable training dataset due to their limited shelf-life. We propose a hybrid dataset generator which utilizes a generative adversarial network (GAN), trained using the publicly available Fruit-360 dataset~\cite{murecsan2017fruit}, to produce self-annotated synthetic fruit images that can be superimposed on to images of real-world fruit grasping scenes for robust image segmentation performance.

We find that (1) GAN-based synthetic dataset improves segmentation performance, even when compared to an alternative hybrid approach that utilizes the Fruit-360 dataset directly, and (2) the proposed synthetic data generator can substantially reduce the labelling time than those of human-involved annotation methods. 
Finally, we evaluate the effect of the proposed dataset generator on a real-world pick-and-place task, and discover that the proposed hybrid dataset dramatically improves labelling and grasping performances.

\section{Related Works}

\subsection{Object detection and segmentation algorithms}

Object detection and segmentation algorithms have been investigated for many robotic applications.
Some traditional techniques, such as using histogram thresholding~\cite{thanammal2014effective} and watershed-based algorithms~\cite{kaur2014image, richard2013fuzzy}, were developed for image segmentation. However, these techniques often necessitate the manual extraction of diverse features meaning that it may be inadequate when dealing with a variety of surrounding environments.

To address these, Deep Learning (DL)-based methods have been applied successfully with the development of Convolutional Neural Networks (CNNs) architectures~\cite{krizhevsky2012imagenet, he2016deep}. However, a large set of well-distributed input images is required to accurately train these algorithms. Even if these CNN-based algorithms have attained state-of-the-art results for object recognition tasks, skewed class distribution and a scarcity of training data remain a severe problem that can cause overfitting and a lack of generalization. Sufficient data acquisition and annotation are challenging since manual image annotation comes with enormous human annotation costs~\cite{taylor2018improving}. To tackle these issues, Segment Anything Model~\cite{kirillov2023segment} has been developed to segment a wide variety of images with a high-quality object masks. However, it demands a substantial amount of computational resources due to its use of Transformer architecture with high-resolution input.
Therefore, we propose a data-efficient way to handle these issues with data augmentation techniques.

Traditional data augmentation algorithms, such as affine and color transformations, have been researched to change certain basic elements of input images~\cite{bjerrum2017smiles}. While these methods have shown significant gains in simple applications, the probability distribution of augmented data input may still be uneven leading to overfitting issues~\cite{perez2017effectiveness}.
In recent years, generative adversarial networks (GANs) and diffusion models have been applied to augment the training dataset, which can generate photo-realistic synthetic images~\cite{zhu2017unpaired, song2019generative}. Diffusion models, however, take longer to generate samples since they rely on a long Markov chain of diffusion steps~\cite{dhariwal2021diffusion}. Therefore, in this work, synthetic objects are produced using GAN to boost the data generation. 

\begin{figure}[t!]
    \vspace{4mm}
    \centering
    \includegraphics[width=0.8\columnwidth]{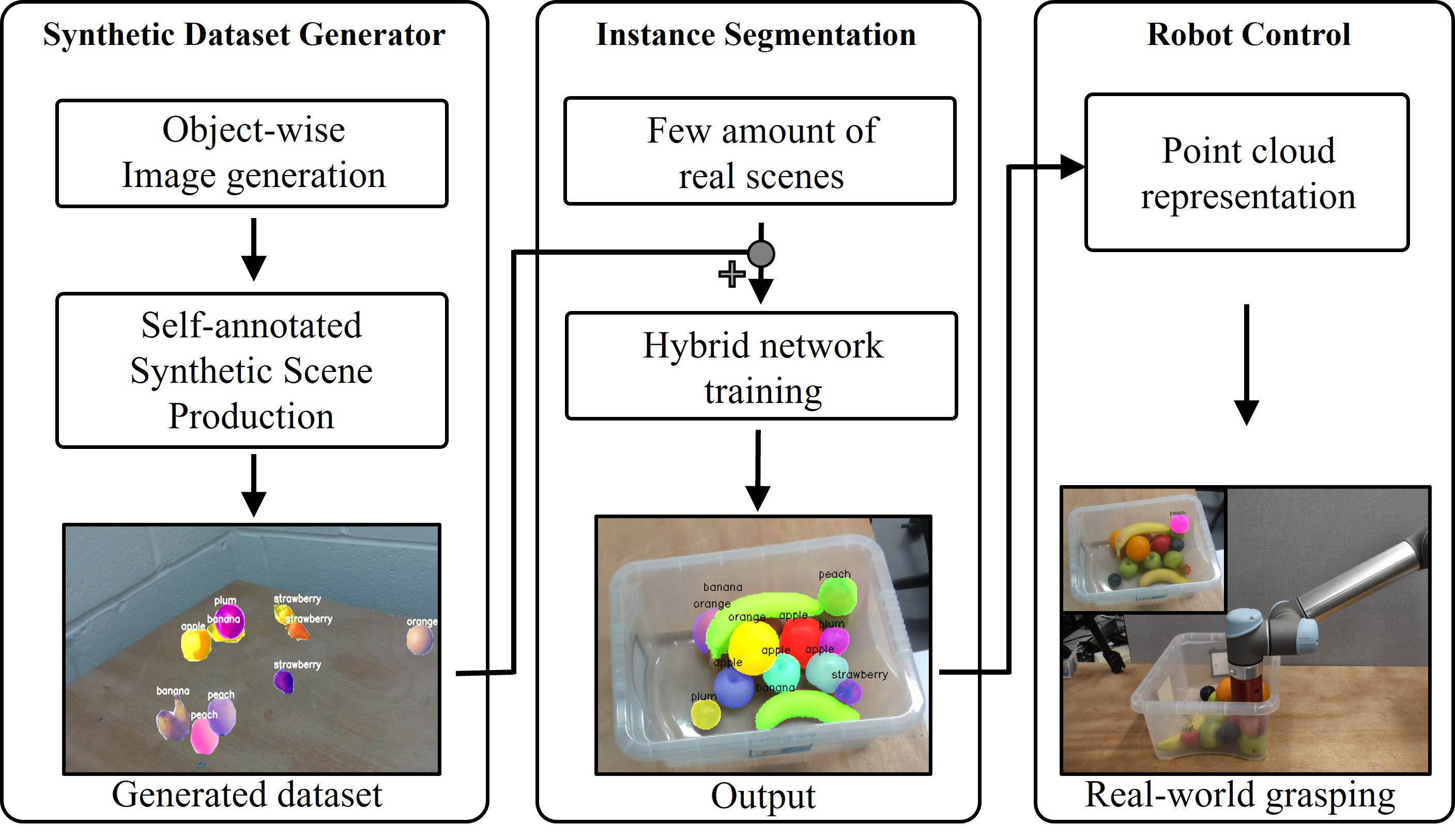}
    \caption{The overall procedure for fruit grasping in clutter.}
    \label{fig:2}
    \vspace{-4mm}
\end{figure}

In terms of GAN, it is challenging to train the GAN algorithm because of its instability and gradient vanishing issue. To address these issues,~\cite{arjovsky2017wasserstein, gulrajani2017improved} developed Wasserstein GAN (WGAN) and its advanced model (WGAN-GP) employing gradient penalty, respectively. The Wasserstein distance is a novel distance metric that indicates the minimum cost to transform the distribution of fake images into the distribution of data input. The Wasserstein distance can exactly estimate the distance gap avoiding the gradient vanishing problem. However, WGAN must use weight clipping to ensure that the weight parameters are 1-Lipschitz function. Furthermore, even if WGAN increases training stability, generated images are still of poor quality and the network sometimes fails to converge. As a result, WGAN-GP is developed to improve the convergence performance by employing gradient penalty (GP) instead of weight clipping. WGAN-GP algorithm is used in this work to produce photo-realistic fruit images that are suitable for this application.

\subsection{Multi-object manipulation}

In real-world applications, multi-object manipulation is required to grasp various instances of same/dissimilar object class in terms of size and shape. To account for shape inconsistency, an object pose estimation algorithm based on a shape-invariant feature representation and graph matching technique was proposed~\cite{sahin2018category}.
However, these approaches make reliable pose estimation challenging in the presence of significant shape and color discrepancy.
As an alternative, a 3D semantic keypoint detection algorithm has been used to identify the target object at the category level to accomplish certain tasks~\cite{qin2020keto}.
While this method has shown promising result for household items, it still requires manual human keypoint specification for the target item.
More recently, a novel canonical representation of intra-class instance candidates is suggested to estimate object pose and size for unseen objects~\cite{wang2019normalized, fu2022category}. The main ideas were (i) a synthetic scene generation method that incorporates real-world table-top scenes and synthetic objects placed on the table in the real image, and (ii) a hybrid dataset to train a model using both synthetic and real-world data, respectively. We construct synthetic scenes in a similar way and integrate them with a little quantity of real-world scenes to produce a hybrid dataset. Furthermore, in this paper, self-annotated synthetic scenes are generated, which can boost the labelled dataset preparation without using any CAD models. 

\section{Methodology}
The overall procedure for multi-fruit grasping, shown in Fig.~\ref{fig:2}, consists of a synthetic dataset generator algorithm to produce table-top synthetic dataset, an instance segmentation algorithm trained with this synthetic data and a limited quantity of real-world images, and real-world grasping demonstration. In this section, we discuss the design of a self-annotated synthetic dataset generator algorithm among these procedures.

\subsection{Object-wise image generation method}
The Fruit-360 dataset is used to implement the GAN-based object-wise image generation. Fruit-360 is a publicly available dataset of fruits and vegetables. Apple, banana, strawberry, orange, peach, and plum are selected as target categories. The number of images and unique instances for the corresponding fruit categories are (6404, 940, 1230, 479, 1722, 1767) and (13, 3, 2, 1, 3, 3), respectively.
While there are enough images and unique instances of apples to train a deep learning model, others are insufficient to do so, especially in terms of the number of unique instances, which makes it difficult to learn intra-class shape and texture disparity. We, therefore, utilize the GAN algorithm to create not only a large amount of photo-realistic dataset but also a diverse shape/texture information.

\begin{figure}[t!]
    \vspace{4mm}
    \centering
    \includegraphics[width=0.95\columnwidth]{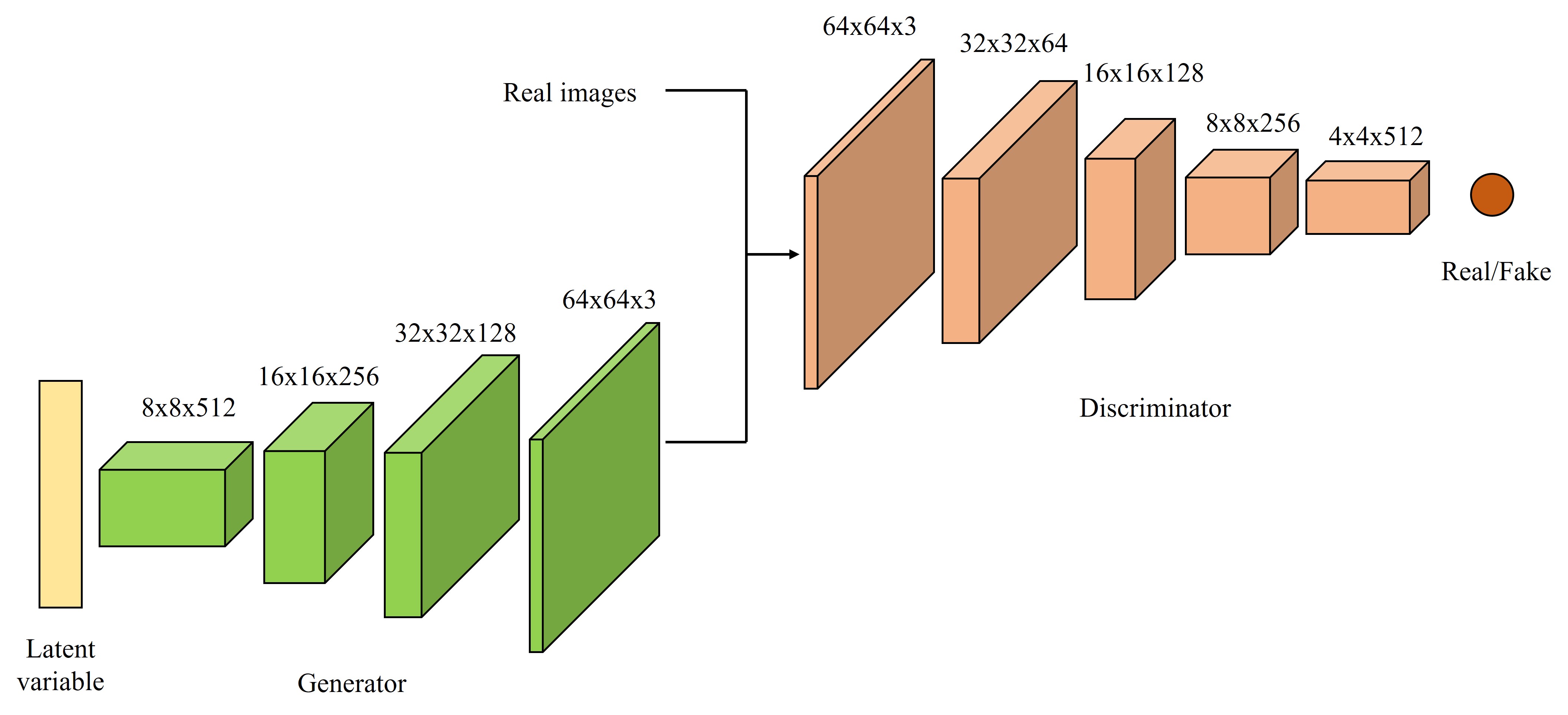}
    \caption{The network architecture of WGAN-GP algorithm.}
    \label{fig:12}
    \vspace{-4mm}
\end{figure}

\begin{figure}[b!]
    \vspace{-4mm}
    \centering
    \includegraphics[width=0.75\columnwidth]{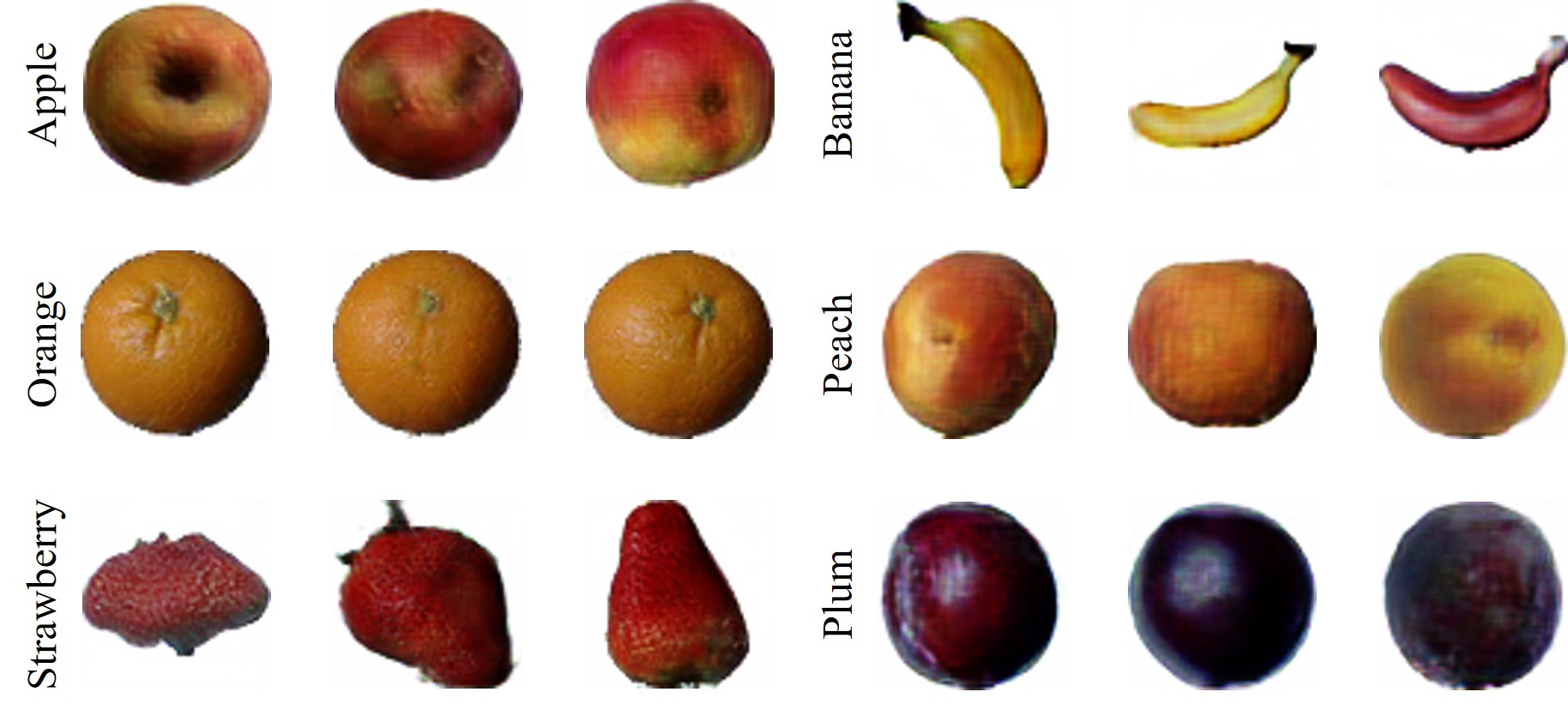}
    \caption{Sample output of generated fruits using WGAN-GP algorithm.}
    \label{fig:3}
    \vspace{-4mm}
\end{figure}

WGAN-GP algorithm is trained for 5,000 epochs and extract 10,000 images for each fruit category. The network architecture of WGAN-GP is illustrated in Fig.~\ref{fig:12}. The entire training is carried out on a single GPU (GeForce RTX 3070, NVIDIA). The generator receives randomly distributed latent space (normal distribution) as an input and the number of latent variables is set to 100, which is one of the most commonly used but efficient value for the latent space dimension~\cite{agustsson2017optimal}. 

ReLU activation function and batch normalization are applied to all convolutional layers of the generator, and LeakyReLU activation function with a negative slope coefficient $\alpha$ = 0.2 is applied to all convolutional layers of the discriminator. The Adam optimizer is chosen to train the WGAN-GP algorithm (learning rate = 1e-4, $\beta_1$ = 0.5, $\beta_2$ = 0.9, and mini-batches of size 32). Table.~\ref{table:1} shows the elapsed time (1) to train the WGAN-GP, and (2) to sample 10,000 images for each fruit category. 

Instead of using the actual image size of Fruit-360 dataset ($100$ x $100$), the input image size is set to $64$ x $64$. More accurate image information is obtained via cropping since redundant background data is eliminated during a generative process.  Fig.~\ref{fig:3} depicts the example output images of generated fruits. This indicates that the generated object can not only be photo-realistic but also include a wide range of shape variations and intra-class information without relying on real-world data.


\begin{figure}[t!]
\vspace{4mm}
    \centering
    \includegraphics[width=0.85\columnwidth]{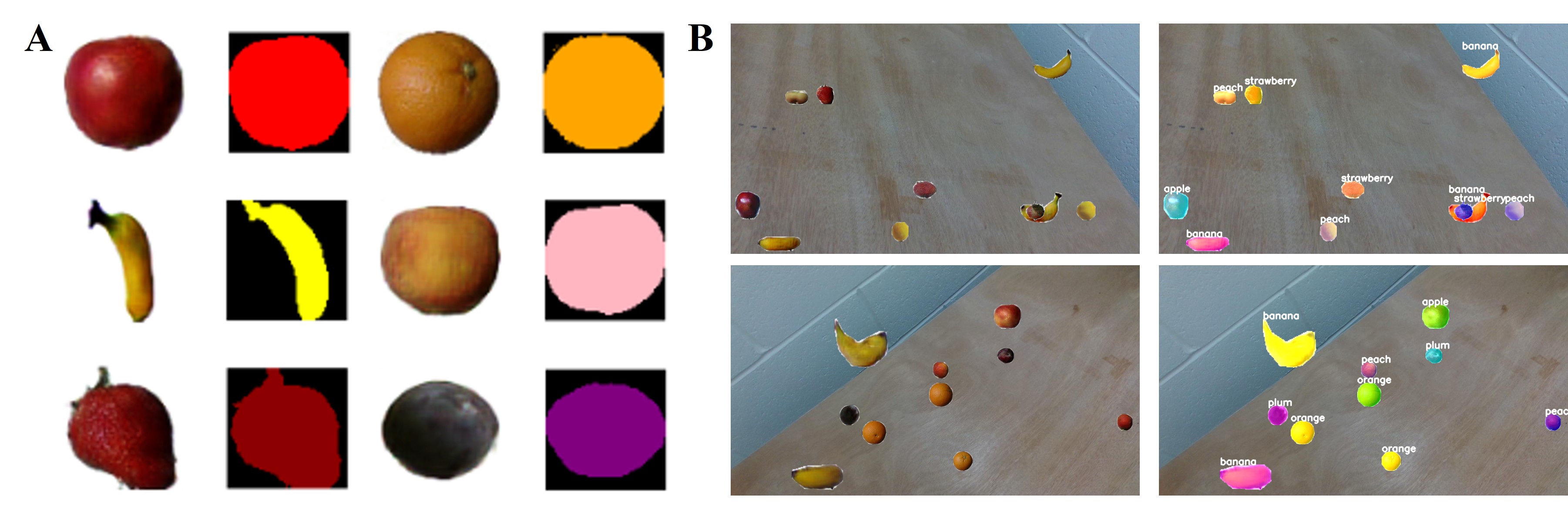}
    \caption{Synthetic scene is generated by randomly pasting object-wise images into the background scenes. \textbf{(A)}: Generated fruit images and segmented pixels representing the target object. \textbf{(B)}: Synthetic scenes with these instances.}
    \label{fig:4}
\vspace{-4mm}
\end{figure}

\begin{table}[b!]
    \vspace{-4mm}
    \centering
    \caption{Elapsed time of WGAN-GP algorithm training and sampling.}
    \resizebox{0.8\columnwidth}{!}{%
    \begin{tabular}{| c | c | c | c | c | c | c |}
        \hline
        Type & Apple & Banana & Strawberry & Orange & Peach & Plum \\
        \hline
        Training (s)  &1404.35 &1395.20 &1411.18 &1397.56 &1389.97 &1433.73 \\
        \hline
        Sampling (s)  &68.30 &70.68 &58.91 &67.29 &68.52 &67.92 \\
        \hline
    \end{tabular}%
    }
    \label{table:1}
    \vspace{-4mm}
\end{table}

\subsection{Self-annotated synthetic scene production algorithm}

Table-top input scenes, containing pixel-level annotations of various fruits and their corresponding background scenes, are necessary to train the instance segmentation algorithm. These scenes are generated using the proposed synthetic scene production algorithm.

Our synthetic scene production algorithm is based on copy and paste (CP) algorithm~\cite{ghiasi2021simple}. A fundamental idea of CP algorithm is to copy instances from one image and paste them randomly into another image using a large scale-jittering method. On the other hand, the proposed algorithm copies the annotated information of an instance from each object-wise image and randomly pastes the selected number of instances into background scenes.

To extract the object itself information from each WGAN-GP output image devoid of background elements, it is essential to perform pixel-wise labelling for each object-wise image. This can be achieved by excluding a range of white values since the background of WGAN-GP output images appears as white. The pixels extracted, which do not fall within this white range, are considered as the segmented components representing the object itself within each object-wise image, as demonstrated in Fig.~\ref{fig:4}(A).
Subsequently, synthetic table-top scenes are generated by placing these annotated objects onto randomly cropped backgrounds, as depicted in Fig.~\ref{fig:4}(B). These table-top synthetic scenes are employed to train the instance segmentation algorithm.




\begin{figure}[t!]
    \vspace{4mm}
    \centering
    \includegraphics[width=0.8\columnwidth]{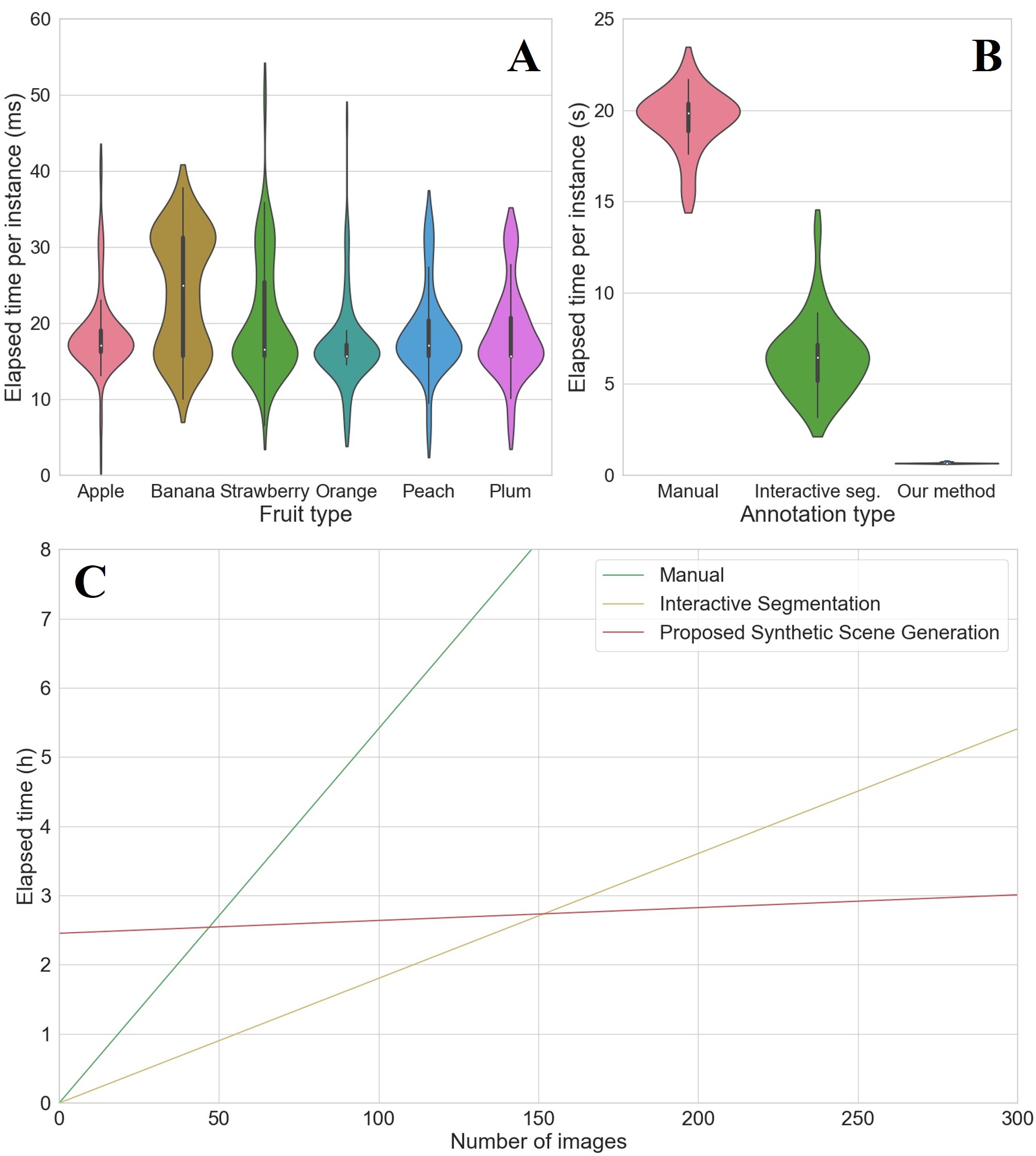}
    \vspace{-1mm}
    \caption{Elapsed time estimation of manual annotation, interactive segmentation, and the proposed synthetic scene generation method. \textbf{(A)}: Self-annotation time per each instance. \textbf{(B)}: Annotation time per each instance for 6-fruit scenes. \textbf{(C)}: Elapsed time estimation by the number of 10-fruit scenes.}
    \label{fig:5}
    \vspace{-4mm}
\end{figure}

\section{Experiments}
The advantage of self-annotated synthetic dataset generator can be proved by comparing (i) the elapsed time to label the training dataset, (ii) the performance of instance segmentation algorithm, and (iii) real-world pick-and-place performance between the proposed algorithm and baseline methods.
\subsection{Annotation elapsed time}
In terms of baselines, human-involved annotation methods are considered: Manual human annotation~\cite{torralba2010labelme} and Click-based interactive segmentation~\cite{sofiiuk2022reviving}. 
The total annotation time of the proposed method is estimated based on the elapsed time of self-annotation of WGAN-GP output images and the subsequent task of copying and pasting these annotated objects onto their corresponding backgrounds. Fig.~\ref{fig:5}(A) shows the self-annotation time of WGAN-GP output for each fruit category and the average elapsed time is 19.64ms.

Fig.~\ref{fig:5}(B), on the other hand, presents the 1-fruit labelling time for 50 multiple fruit-included scenes, comparing human-involved annotation methods with the proposed algorithm.
Click-based interactive segmentation can substantially accelerate the annotation particularly for uncluttered objects than manual annotation method, however, labelling for highly cluttered objects is relatively decelerated since multiple same-type instances are sometimes recognized as a single instance. Based on these results, overall dataset preparation time are estimated by the number of 10-fruit scenes, which are used as an actual training dataset of instance segmentation algorithm. In case of the proposed method, elapsed time is not zero when the number of 10-fruit scene is zero since WGAN-GP training and image sampling are required as a prior condition. This result shows that the proposed synthetic dataset generator can significantly diminish not only the human-annotation time but also the absolute wall-clock dataset preparation time when a large size of training dataset becomes necessary.

\begin{table}[b!]
    \vspace{-4mm}
    \centering
    \caption{Average precision and recall of instance segmentation algorithm with the Real-only, Syn-only, and CP-only datasets.}
    \resizebox{0.7\columnwidth}{!}{%
    \begin{tabular}{| c | l | c | c | c | c | c | c | c | c |}
        \hline
        \multirow{2}{*}{\bf Dataset} & \multirow{2}{*}{\bf Metric (\%)} & \multicolumn{8}{c|}{\bf Number of images} \\
        \cline{3-10}
        & & 50 & 100 & 150 & 200 & 300 & 400 & 500 & 600 \\
        \hline
        \multirow{6}{*}{Syn-only} & AP@0.5 &61.2 &69.3 &67.8 &77.7 &79.6 &\bf 80.6 &76.5 &71.0 \\
                                        & AP@0.75 &59.0 &67.3 &67.9 &75.0 &77.5 &\bf 77.8 &74.0 &68.4 \\ 
                                        & AP@[0.5:0.95] &50.4 &60.1 &60.7 &68.1 &70.5 &\bf 72.5 &65.9 &61.7 \\
        \cline{2-10}
                                        & AR@0.5 &89.8 &85.2 &87.4 &89.2 &89.6 &\bf 89.9 &85.8 &83.8 \\ 
                                        & AR@0.75 &78.4 &78.7 &81.1 &83.2 &\bf 84.9 &84.6 &80.7 &76.2 \\ 
                                        & AR@[0.5:0.95] &65.1 &67.7 &68.2 &73.4 &74.9 &\bf 75.1 &73.1 &67.1 \\ 
        \hline
        \multirow{6}{*}{CP-only} & AP@0.5 &60.1 &61.9 &65.2 &68.4 &71.9 &\bf 72.5 &69.4 &67.7 \\
                                        & AP@0.75 &56.8 &60.2 &63.8 &66.8 &70.3 &\bf 70.8 &67.5 &66.0 \\ 
                                        & AP@[0.5:0.95] &49.8 &55.0 &58.6 &61.5 &65.0 &\bf 65.6 &62.9 &61.0 \\
        \cline{2-10}
                                        & AR@0.5 &89.3 &86.1 &84.5 &85.0 &\bf 86.0 &85.8 &82.2 &84.7 \\ 
                                        & AR@0.75 &76.0 &80.8 &80.5 &81.6 &82.5 &\bf 83.4 &79.8 &80.2 \\ 
                                        & AR@[0.5:0.95] &66.2 &69.8 &69.6 &71.4 &72.9 &\bf 73.4 &71.0 &70.5 \\
        \hline
        \multirow{6}{*}{Real-only} & AP@0.5 &84.0 &83.7 &85.0 &\bf 85.1 &- &- &- &- \\
                                        & AP@0.75 &81.5 &81.4 &82.5 &\bf 82.5 &- &- &- &- \\
                                        & AP@[0.5:0.95] &74.2 &75.7 &77.7 &\bf 78.4 &- &- &- &- \\
        \cline{2-10}
                                        & AR@0.5 &\bf 92.8 &92.1 &92.2 &92.6 &- &- &- &- \\
                                        & AR@0.75 &88.1 &87.6 &89.2 &\bf 89.4 &- &- &- &- \\
                                        & AR@[0.5:0.95] &76.7 &79.7 &81.9 &\bf 82.9 &- &- &- &- \\
        \hline
    \end{tabular}
    }
    \label{table:2}
    \vspace{-3mm}
\end{table}

\subsection{Instance segmentation performance}
PyTorch-based implementation for Mask R-CNN~\cite{massa2018mrcnn} is employed. ResNet-50-FPN backbone, which is pre-trained on MS COCO~\cite{lin2014microsoft}, is applied to improve the performance of an instance segmentation technique.
Table-top synthetic scenes are refined using traditional data augmentation algorithms to mitigate overfitting by simulating realistic environmental conditions using the albumentation library~\cite{bjerrum2017smiles, buslaev2020albumentations}.
Input images are scaled to $1280$ x $720$ with preserving aspect ratio using zero-padding if necessary.
The results of instance segmentation algorithms are presented using the standard evaluation metrics of average precision (AP) and recall (AR). Intersection over Union (IoU) is an important value to assess the AP and AR. It is calculated using overall regions and intersections of bounding boxes or masks. Given the wide range of cluster sizes and shapes, IoU values indicate a degree of coincidence in terms of pixels between actual and predicted target instances.
Fruit-360 dataset (CP-only), which only apply the proposed synthetic scene generation method without using GAN algorithm, is also compared with synthetic dataset (Syn-only) as a baseline like a real-world scene-based dataset (real-only). As a result, the network performances trained with real-only, CP-only, and Syn-only datasets are measured in terms of AP and AR over IoU thresholds. To evaluate the capacity to segment the region of each detected instance, AP at (1) 0.5 IoU threshold, (2) 0.75 IoU threshold, and (3) over IoU thresholds in the range from 0.5 to 0.95 with an interval of 0.05 are presented. To evaluate the performance of object detection, AR based on bounding boxes at the same IoU thresholds is reported. 

\begin{figure}[t!]
\begin{center}
\vspace{4mm}
\includegraphics[width=0.8\columnwidth]{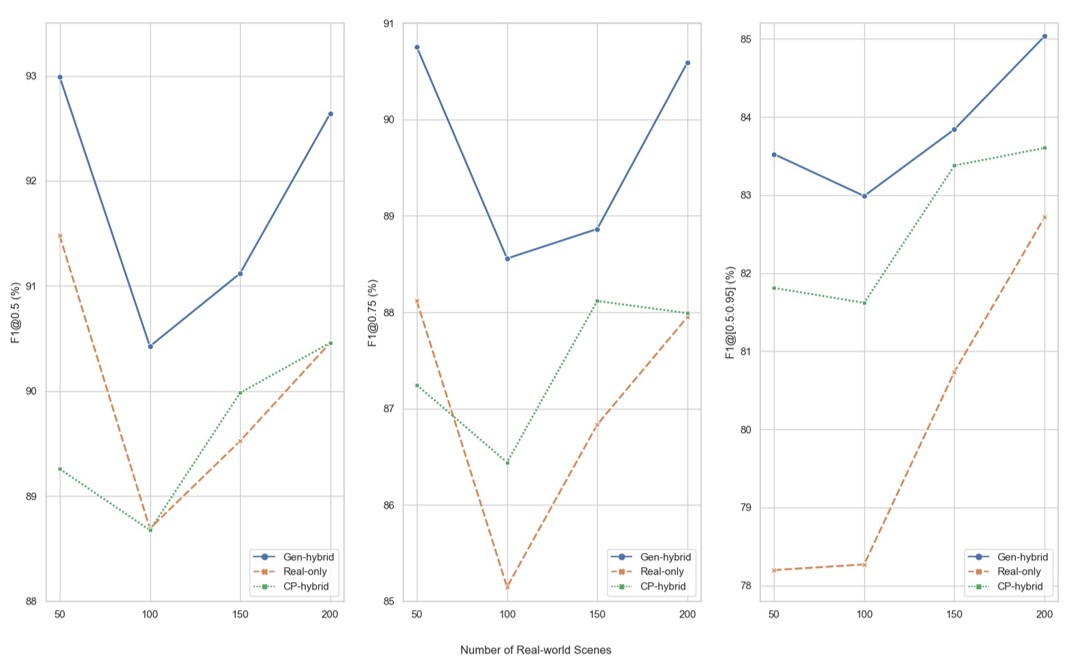}
\end{center}
\vspace{-4mm}
\caption{F1 score, which is a harmonic mean of average precision and recall, of the best real-only, CP-hybrid, and Gen-hybrid networks.}\label{fig:6}
\vspace{-4mm}
\end{figure}

\begin{figure}[b!]
\vspace{-4mm}
\begin{center}
\includegraphics[width=0.75\columnwidth]{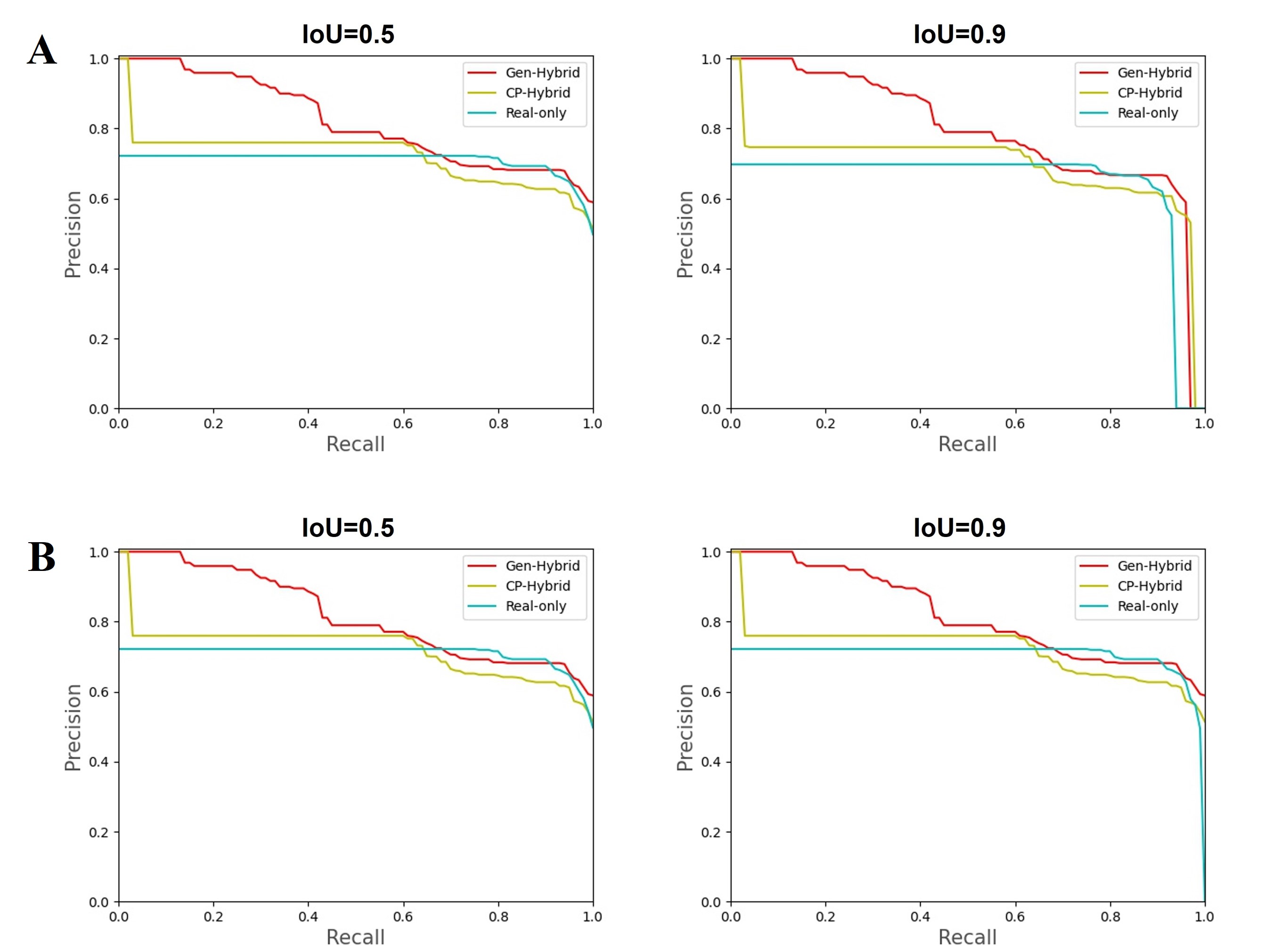}
\end{center}
\caption{The proposed hybrid network (Gen-hybrid) is compared with the baseline networks in terms of \textbf{(A)}: Bounding box-based PR curves and \textbf{(B)}: Mask-based PR curves.}\label{fig:7}
\vspace{-4mm}
\end{figure}

Table.~\ref{table:2} shows the evaluation results of segmentation algorithms trained with real-only, CP-only, and Syn-only datasets. The test set consists of 200 manually annotated real table-top scenes with different numbers and types of fruits. As for Syn-only and CP-only datasets, the performance of Syn-only dataset outperforms that of CP-only in general, especially in terms of AP. However, Syn-only network is not comparable to the real-only network performance. The main reason may be the self-annotating method. Self-annotated objects are not as accurate as human-annotated items since all the non-white pixels are segmented as an object according to the proposed self-annotating algorithm.
Furthermore, regardless of fruit category, all fruit images in the Fruit-360 dataset have a fixed resolution of $100$ x $100$. Even if large scale-jittering is used, the image quality is not comparable to that of a real-world scene.

To overcome the limitation of network training only with synthetic datasets, the hybrid dataset is proposed consisting of synthetic and real-world scenes.
We found the best hybrid networks, both for synthetic and Fruit-360 datasets, in terms of AP and AR for different numbers of real images (e.g. 50, 100, 150, and 200). The optimal network for each number of real images is found via comparing the performance of networks trained on hybrid data across a range of synthetic images, from 50 to 400 (e.g. 50, 100, 150, 200, 300, and 400). As illustrated in Fig.~\ref{fig:6}, the F1 score of network trained with synthetic and real-world data (Gen-hybrid) outperforms not only the real-only network but also the network trained on Fruit-360 and real-world input (CP-hybrid). The performance of Gen-hybrid network is significantly improved when the number of real image is limited (i.e., 50). 

To evaluate the performance of real-only, Gen-hybrid, and CP-hybrid networks in detail, the precision-recall (PR) curve of these networks, when the number of real-world scenes is 200, are illustrated in Fig.~\ref{fig:7}. This result supports the fact that the performance of Gen-hybrid network substantially outperforms the baseline networks in general. While synthetic data slightly hinder the segmentation due to a lack of generated image quality in some recall ranges, synthetic data considerably aids the segmentation algorithm to gain the higher precision, especially in the range of recall from 0.0 to 0.6.

\begin{figure}[t!]
\vspace{4mm}
\begin{center}
\includegraphics[width=0.6\columnwidth]{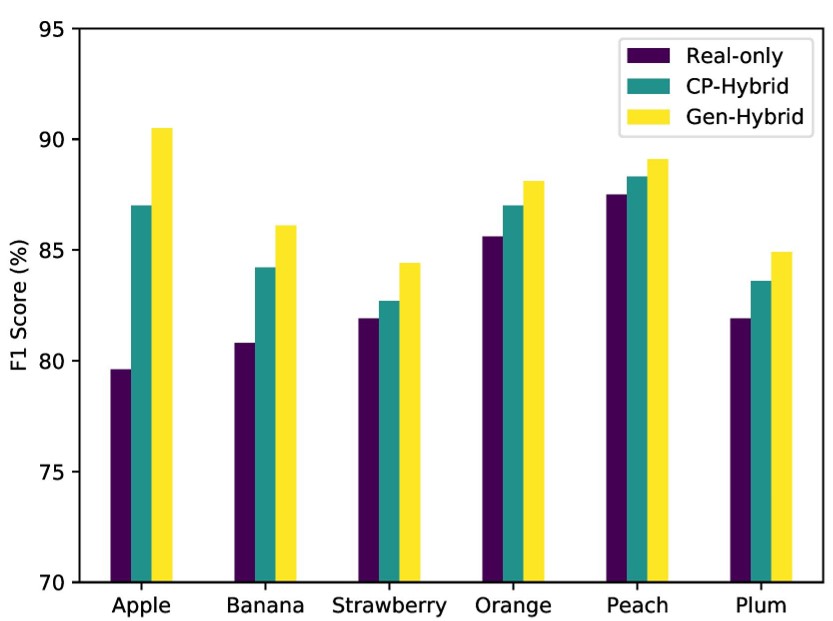}
\end{center}
\caption{The performance comparison between real-only, CP-hybrid, and Gen-hybrid networks in terms of object category.}\label{fig:8}
\vspace{-6mm}
\end{figure}

Then, we compute F1 score (over IoU thresholds in the range from 0.5 to 0.95 with an interval of 0.05) of real-only, Gen-hybrid, and CP-hybrid networks in terms of fruit category, when the number of real-world scene is 200, as in Fig.~\ref{fig:8}. The number of images and unique instances for each category in Fruit-360 dataset significantly affect the perception performance when comparing the gap between real-only and CP-hybrid. More interestingly, Gen-hybrid outperforms CP-hybrid in a similar way since GAN network is better trained as the number of images and unique instances increase. For instance, the case of apple exhibits the most considerable improvement since the amounts of images and unique instances are significantly larger than those of other fruits. It demonstrates that GAN-based synthetic dataset covers more diverse shape and texture information rather than Fruit-360 dataset itself.





\begin{table}[b!]
    \vspace{-4mm}
    \centering
    \caption{Performance of point cloud segmentation methods.}
    \resizebox{0.6\columnwidth}{!}{%
    \begin{tabular}{| c | c | c | c |}
        \hline
        \textbf{Methods} & \textbf{AP (\%)} & \textbf{AR (\%)} & \textbf{F1 score(\%)} \\
        \hline
        K-means     &57.8 &89.2 &70.1 \\
        \hline
        DBSCAN      &58.1 &89.4 &70.4 \\
        \hline
        Gen-Hybrid  &91.6 &92.6 &92.1 \\
        \hline
    \end{tabular}
    }
    \vspace{-4mm}
    \label{table:3}
\end{table}

\begin{figure}[t!]
\vspace{4mm}
\begin{center}
\includegraphics[width=0.8\columnwidth]{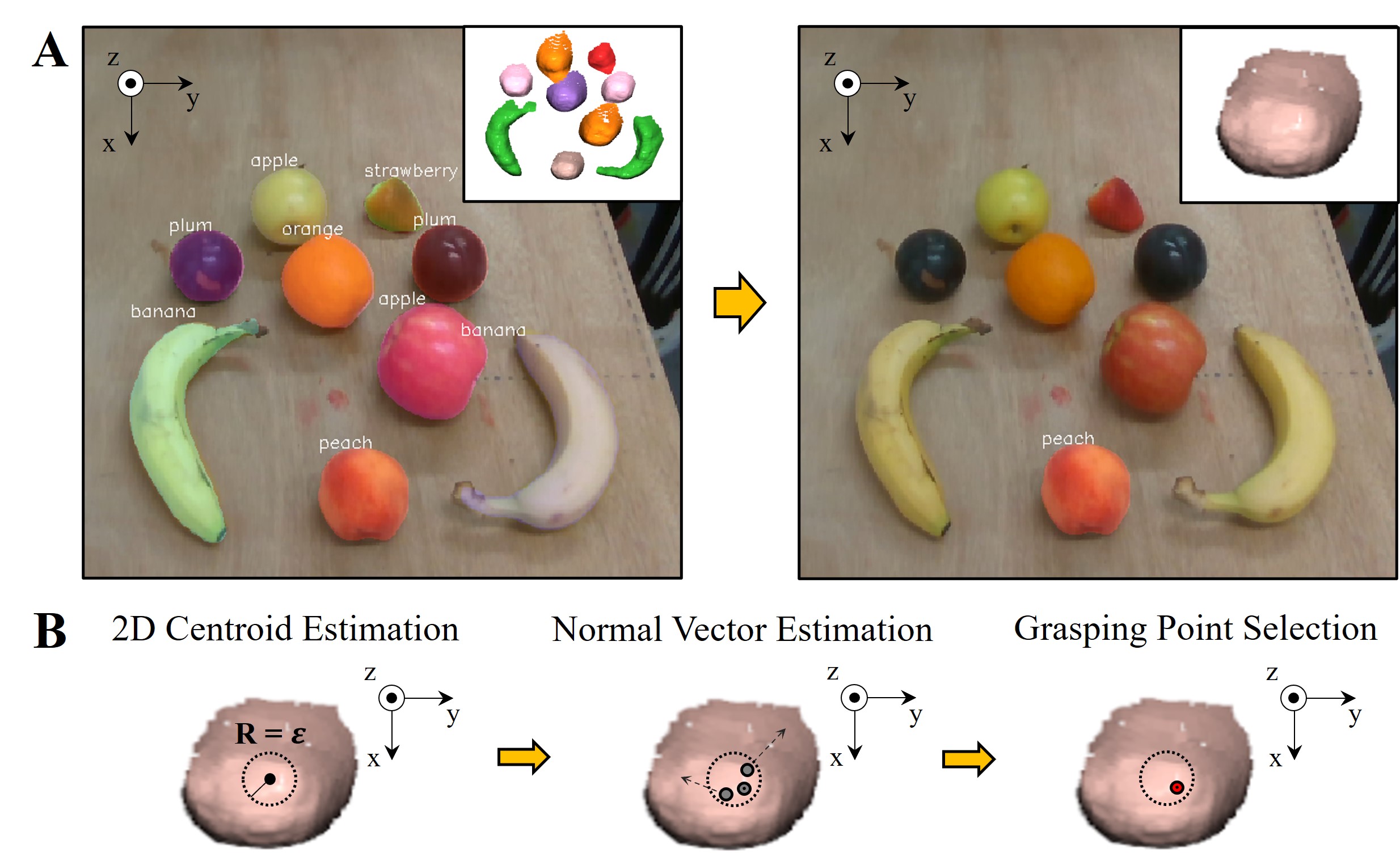}
\end{center}
\caption{\textbf{(A)}: Target selection result, and \textbf{(B)}: Grasping point selection method for the target instance.}\label{fig:10}
\vspace{-4mm}
\end{figure}

\subsection{Real-world pick-and-place operation}
\subsubsection{Object localization}
It is necessary to evaluate whether the proposed segmentation algorithm performs well or not in terms of point cloud to perform the real grasping experiment. Then, we compared the performance of the proposed algorithm with Euclidean clustering-based methods (K-means clustering~\cite{duda1973pattern} and DBSCAN~\cite{ester1996density}) with 50 test sets. Euclidean clustering-based methods are chosen since other point cloud segmentation models require prior knowledge of CAD models or topological representations at least during the training stage. Due to the fact that these baseline methods can only get the cluster information, these methods are compared with the proposed algorithm in terms of overall AP and AR, meaning that 6 different types of fruits are all assigned as a single category for a fair comparison, as shown in Table.~\ref{table:3}. In terms of AR, they are comparable with Gen-hybrid network since some pre-processing techniques are applied such as random sample consensus (RANSAC). However, despite the usage of these pre-processing functions, there is still 33.6\% performance gap on average between Gen-hybrid network and baseline methods in terms of AP due to a large amount of false positive information, indicating that the segmented points of baseline methods are frequently not the actual target objects. The proposed algorithm has achieved a high F1 score of 92.1\% in terms of point cloud segmentation.

\subsubsection{Target selection}
In addition, the most graspable object should be chosen for the highly cluttered scenarios before finding the optimal grasping point. After applying the instance segmentation algorithm, graspable objects are detected based on Mask R-CNN confidence score, representing the likelihood that the prediction of instance segmentation algorithm is correct. The highest-scored instance is selected as the most graspable object as shown in Fig.~\ref{fig:10}(A).

In terms of grasping pose, it is not fully defined as we do not include any CAD models even for the training stage. Rather than directly associating the grasping pose with the overall 3D model of the instance, the desired grasping point can be defined, allowing the suction gripper to approach and grasp the target instance. Therefore, a geometry-based method is applied to acquire the optimal grasping point.
A perpendicular approach of an end-effector is required since other items can collide with the suction gripper during the grasping procedure.
Furthermore, a grasping point should be placed adjacent to the centroid of the target object. In this case, 2D $x$-$y$ centroid is considered since an end-effector approaches to the target instance vertically. As we proved the point cloud segmentation performance of the proposed method in Table.~\ref{table:3}, 2D centroid of a target object can be predicted by the calculation of the average $x$-$y$ position in the map of a point cloud. Then, a normal vector of each point within a certain region, which has a radius ($R$ = 10mm) and its center is the average $x$-$y$ position of object points, is calculated. The point with the closest normal vector to the perpendicular unit vector in the world coordinate system, as shown in Fig.~\ref{fig:10}(B), is chosen as the optimal grasping point among the candidates near the 2D centroid.

\begin{figure}[t!]
\vspace{4mm}
\begin{center}
\includegraphics[width=0.9\columnwidth]{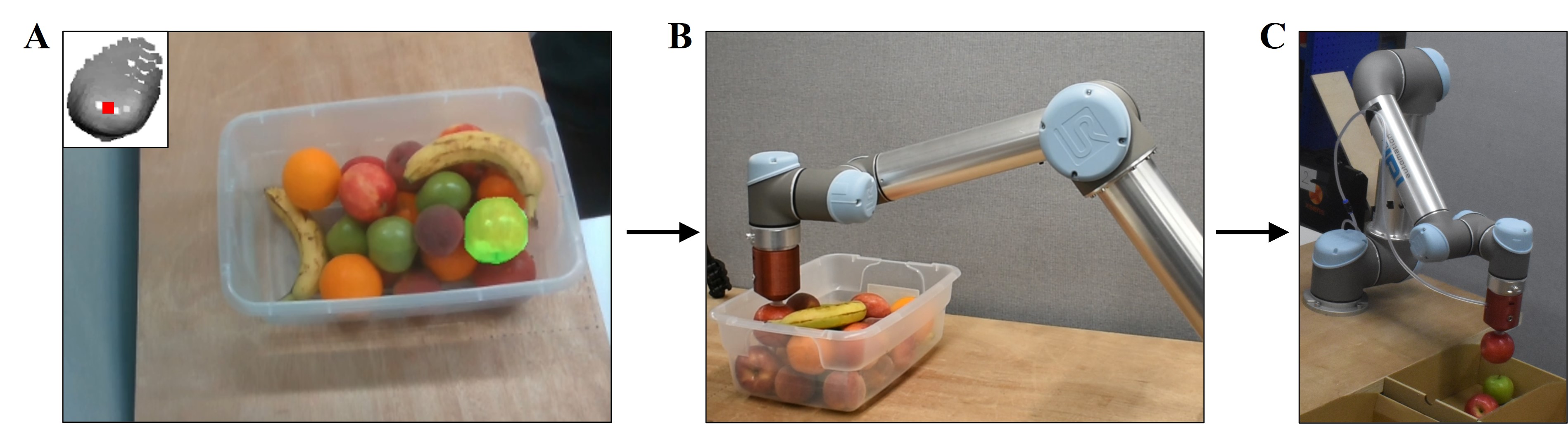}
\end{center}
\vspace{-2mm}
\caption{Real-world pick-and-place demonstration procedure.}\label{fig:11} 
\vspace{-4mm}
\end{figure}

\subsubsection{Real-world demonstration}
Pick-and-place experiments in cluttered settings are carried out to demonstrate the accuracy of (i) the extracted label of target object (Labelling) and (ii) segmented pixels of the target object (Grasping), as shown in Fig.~\ref{fig:11}. The objective of the pick-and-place operation is to identify and pick up the most easily graspable object and correctly place it into the corresponding labelled box for object sorting. Labelling and grasping success rates are provided in Table.~\ref{table:4}. UR5 robotic arm and the customized suction gripper are used. To assess the robustness of the proposed algorithm, different instances of each fruit type are considered. In terms of the experiment setup, highly cluttered scenarios are considered with 12 stacked real-world fruits in a box. The condition of grasp failure is determined by the grasping and holding ability until the robotic arm gets to the target box. On the other hand, the state of labelling success is decided by the extracted class label of the most graspable object. Average labelling and grasping success rate of 98.9 and 70 percents are achieved, respectively.
In terms of grasping success rate, it generally follows the trend of an evaluation result in Fig.~\ref{fig:8}. The main reasons of grasping failure are following: (i) grasping point is occasionally imprecise in cases of occlusion since the experiment considers highly cluttered scenarios, (ii) even if the target is accurately defined, it is difficult to grasp it securely with the suction gripper when its surface is irregular or slippery, (iii) the target is sometimes soft making it vulnerable to be damaged, especially in case of strawberry, and (iv) the diameter of the suction cup may not fit to some small and light fruits. On the other hand, labelling failure is mainly caused by a similar appearance and occlusion. The proposed algorithm sometimes confuses between (i) red apple and peach or (ii) peach and plum. Other cases are primarily related to the occlusion indicating that the target is partially or fully occluded by a box or other fruits.

\section{Conclusion} 
Robotic applications for multi-object grasping have been developed with CNN-based object localization algorithms. However, training data acquisition without access to CAD models are still challenging. We utilized WGAN-GP algorithm with self-annotation method to acquire a synthetic dataset to train instance segmentation algorithm. 
The proposed synthetic scene production algorithm significantly boosts the training dataset preparation rather than human-involved annotation methods.
Our object recognition and localization methods can be applied without any 3D models even in highly cluttered scenarios. We show that synthetic scenes can be used as training input with few amount of real data to significantly improve the performance of instance segmentation. 
Furthermore, while the proposed algorithm only considers RGB information, we demonstrate that it considerably outperforms traditional point cloud segmentation techniques. Real-world sorting experiments are successfully implemented to ensure the viability of the proposed network. 
Further study could investigate deep learning-based 6D pose estimation without access to instance models for grasping with a robot hand, as well as more refined synthetic scene generation algorithms.

\begin{table}[t!]
    \vspace{4mm}
    \centering
    \caption{Sorting success rate for highly cluttered settings.}
    \resizebox{0.7\columnwidth}{!}{%
    \begin{tabular}{|c|c|c|c|c|c|c|c|c|}
        \hline
        \multirow{2}{*}{\bf Items} & \multirow{2}{*}{\bf Instance} & \multicolumn{2}{c|}{\bf Real-only} & \multicolumn{2}{c|}{\bf CP-Hybrid} & \multicolumn{2}{c|}{\bf Gen-Hybrid} \\
        \cline{3-8}
         & & Label & Grasp & Label & Grasp & Label & Grasp \\
        \hline

        Apple      &12 &27/30 &17/30 &28/30 &20/30 &\textbf{29/30} &\textbf{24/30} \\
        Banana     &6  &27/30 &17/30 &29/30 &18/30 &\textbf{30/30} &\textbf{21/30} \\
        Strawberry &12 &30/30 &12/30 &30/30 &15/30 &\textbf{30/30} &\textbf{17/30} \\
        Orange     &6  &26/30 &15/30 &28/30 &18/30 &\textbf{30/30} &\textbf{22/30} \\
        Peach      &8  &27/30 &15/30 &28/30 &18/30 &\textbf{29/30} &\textbf{20/30} \\
        Plum       &12 &29/30 &17/30 &30/30 &19/30 &\textbf{30/30} &\textbf{22/30} \\
        \hline
        Total    &56 &92.2\% &51.2\% &96.1\% &60\% &\textbf{98.9\%} &\textbf{70\%} \\
        \hline
    \end{tabular}
    }
    \label{table:4}
    \vspace{-4mm}
\end{table}

\addtolength{\textheight}{-1cm} 

\bibliographystyle{IEEEtran}
\bibliography{references}

\end{document}